\documentclass[journal]{IEEEtran}
\pdfoutput=1

\ifCLASSINFOpdf
\usepackage[pdftex]{graphicx}
\else
\usepackage[dvips]{graphicx}
\fi

\usepackage{cite}
\usepackage[super]{nth}
\usepackage{subfig}
\usepackage{amsmath}
\usepackage{booktabs}
\usepackage[export]{adjustbox}
\usepackage{url}

\begin{document}

\title{SuSketch: Surrogate Models of Gameplay as a Design Assistant}
\author{Panagiotis Migkotzidis and Antonios Liapis,~\IEEEmembership{Member,~IEEE}
\thanks{All authors are with the Institute of Digital Games, University of Malta, Msida, MSD 2080, Malta}
}

\markboth{IEEE Transactions on Games}{SuSketch: Surrogate Models of Gameplay as a Design Assistant}

\maketitle

\begin{abstract}
This paper introduces SuSketch, a design tool for first person shooter levels. SuSketch provides the designer with gameplay predictions for two competing players of specific character classes. The interface allows the designer to work side-by-side with an artificially intelligent creator and to receive varied types of feedback such as path information,  predicted balance between players in a complete playthrough, or a predicted heatmap of the locations of player deaths. The system also proactively designs alternatives to the level and class pairing, and presents them to the designer as suggestions that improve the predicted balance of the game. SuSketch offers a new way of integrating machine learning into mixed-initiative co-creation tools, as a surrogate of human play trained on a large corpus of artificial playtraces. A user study with 16 game developers indicated that the tool was easy to use, but also highlighted a need to make SuSketch more accessible and more explainable.
\end{abstract}


\IEEEpeerreviewmaketitle

\section{Introduction}\label{sec:introduction}

\IEEEPARstart{R}{ecent} advances in artificial intelligence (AI) have led to breakthroughs in automating a broad set of tasks, including that of game design. While procedural content generation (PCG) has a long history in the game industry, the last decade has seen a strong academic interest in enriching generative algorithms with sophisticated AI in order to speed-up generation, improve the quality of content, or explore new types of content that can be generated \cite{liapis2020tenyears,yannakakis2016panorama}. A large part of this academic movement has also focused on the controllability of the algorithms, and a number of design tools have been proposed to allow human designers and AI to co-create content in a mixed-initiative manner \cite{liapis2016mixedinitiative}. Sentient Sketchbook \cite{liapis2013sketchbook}, for example, generated and presented suggestions in real-time while the designer was editing the level. The interface allowed the user to review a number of calculated metrics regarding their level (such as number of resources, or degree of exploration between the two players' bases), and compare their level with the generated suggestions in terms of these metrics. As another example, Morai Maker \cite{guzdial2019moraimaker} allows the user and the AI to take turns editing a platformer level for \emph{Super Mario Bros.} (Nintendo, 1985). Besides the turn-taking interaction \cite{novick1997mixedinitiative}, Morai Maker also handles generation differently from most AI-assisted game design tools. Morai Maker leverages machine learning by training a model on existing Super Mario Bros. levels in order to detect local patterns between tiles. For generation, the tool uses the recent PCG via machine learning (PCGML) paradigm \cite{summerville2018pcgml} to adjust the designed levels in order to better match the level design patterns found in the corpus.

This paper presents SuSketch, a design tool which takes best practices from the literature and exploits PCGML in a different way, using machine learned models as \emph{surrogates of gameplay} rather than as a collection of level patterns. SuSketch builds on the extensive work of Karavolos \emph{et al.} \cite{karavolos2018pairing,karavolos2018surrogate,karavolos2019multifaceted} on mapping first person shooter (FPS) levels and the competing players' classes to gameplay outcomes such as the duration and winner of the match. Using the extensive corpus collected by this work, SuSketch uses surrogate models to inform a designer on the predicted gameplay outcomes of their work-in-progress. This real-time feedback allows the designer to tweak parts of the level or the character classes assigned to each player, in order to ensure a balanced matchup when the level is finally played. Moreover, SuSketch generates suggestions for the character classes and powerup placement in the level, towards countering any (predicted) advantage a player may have over the other. SuSketch is the first instance where AI assistance---in the form of both visualizations and generated suggestions---is based on surrogate models of gameplay, which consider both the level layout and the game's rules (in the form of character classes) in a multi-faceted fashion \cite{karavolos2019multifaceted}. While current mixed-initiative tools estimate gameplay based on pathfinding information \cite{liapis2013sketchbook} or computationally heavy simulations \cite{shaker2013ropossum}, SuSketch can predict, visualize, and account for the spatial, temporal and end-game outcomes of a playthrough while providing immediate feedback to the designer. 

This paper extends the work of Karavolos \emph{et al.}~\cite{karavolos2018pairing} in significant ways, by incorporating the surrogate models into a level editing tool and using them to generate a new type of suggestion focusing solely on powerup placement. While the dataset and surrogate models of kill ratio and match duration originate from earlier work \cite{karavolos2019multifaceted}, this work introduces new gameplay metrics that are multi-dimensional: namely, the progression of the dramatic arc and combat pacing over time and the spatial distribution of player deaths. The user study in this paper also investigates usability issues of the tool as well as how users react to the novel additions of real-time gameplay predictions and the generated level or class suggestions.

\section{Related Work}\label{sec:related}

This section discusses the state-of-the-art in the relevant fields that SuSketch builds upon.

\subsection{Mixed-Initiative Design}\label{sec:related_mixedinit}

Mixed-initiative design tools are computer-aided design (CAD) software where both the human and the computer proactively make contributions to the problem solution \cite{yannakakis2014micc}. The computational initiative in CAD software has a rich history, dating back to the concept of human-computer symbiosis in the 1960s \cite{licklider1960mancomputer}. Novick and Sutton \cite{novick1997mixedinitiative} treat mixed-initiative interaction as analogous to a conversation and identify components of initiative as: (a) task initiative (deciding the topic of the conversation), (b) speaker initiative (deciding when each actor can speak), and (c) outcome initiative (deciding how and when the discussion is resolved). A plethora of mixed-initiative tools have been introduced for engineering \cite{holland2003principles}, art \cite{davis2016empirically}, and design \cite{urban5}. This section focuses on mixed-initiative tools in the field of game design, where domain-specific expert knowledge is integrated into the AI to support novices throughout the design process.

Sentient Sketchbook \cite{liapis2013sketchbook} supports the design of map sketches via a tile canvas interface, providing real-time evaluations of game level quality via heuristics and visual aids. The tool assists the level designer as it automatically tests maps for playability constraints, calculates and displays navigable paths, evaluates the map on gameplay properties and adds details to the coarse map sketch.  Similar to SuSketch, Sentient Sketchbook uses a map recommendation system via computer-generated suggestions evolved from the users’ own creations. SuSketch extends the paradigm of Sentient Sketchbook by evaluating the level content through surrogate models of game simulations rather than static pathfiding. The same paradigm was taken to another direction by the Evolutionary Dungeon Designer (EDD) \cite{alvarez2018fostering}, which allows the user to design a dungeon in a piece-meal fashion via a two-step process: at the macro-level where cell connectivity is determined and at the micro-level where the tiles of each cell are determined. EDD can indicate design patterns in work-in-progress levels, and provides suggestions based on symmetry, patterns or novelty. 

While Sentient Sketchbook and EDD use fitness functions primarily based on path distances to generate suggestions for a designer, playability can be tested in many different ways. Tanagra \cite{smith2010tanagra} provides platformer level suggestions in real-time which are guaranteed to be playable based on constraints on how high an avatar can jump. Tanagra uses constrained programming to achieve this, which is also used by Butler \emph{et al.} to ensure playability of generated puzzles for the math game Refraction \cite{butler2013puzzle}. Refraction allows the designer to decide which operations are available to the player, and the generator ensures via constraints that any generated level can be solved with these operations. While puzzle games or top-down strategy levels can sufficiently be tested for playability in a static fashion from the initial game-state, other genres such as physics games require actual playthroughs. In this vein, Ropossum \cite{shaker2013ropossum} allows designers to test their \textit{Cut The Rope} (Chillingo, 2010) levels by playing them on their own or via an AI agent. Although computationally expensive, such AI playtests offer an animation of how the level is solved but also show which actions are required. Ropossum can provide playable suggestions by modifying some of the elements in the level, although the time needed to compute these makes it unwieldy as a design tool. SuSketch attempts to minimize this feedback delay by replacing simulations with a surrogate model that predicts the simulation's outcomes.

Finally, it is worth noting that not all mixed-initiative tools pertain to level design. For instance, Mek allows designers to define a set of spatial interaction rules by directly ``painting'' the rules' effects \cite{volkovas2019mek}. Machado \emph{et al.} \cite{machado2016mechanics} introduced a mechanics recommendation system based on similarity to existing games in the corpus of GVGAI \cite{perez2019gvgaibook} games. The mechanics recommender was integrated into Cicero \cite{machado2018ai} which allowed users to design levels and mechanics, providing extensive visualizations and analytics. Since the games produced in Cicero are still in the GVGAI framework, the tool allows designers to pick any of the AI agents available and simulate the game, providing visualizations of e.g. the trajectories of enemies, events that occurred, and a frame-by-frame replay of the game. The mechanics recommender was extended in Pitako \cite{machado2019pitako}, which recommends mechanics and dynamics to the user on the game they are making. Through a confidence score, designers can apply suggestions that offer common mechanics for their game, or explore and try different possibilities. SuSketch follows a similar multi-facet recommendation system, orchestrating the level design and rule design facet \cite{liapis2019orchestrating} by providing suggestions on finding the most balanced powerup placement as well as class matchup. 

\subsection{Machine Learning for Procedural Content Generation}\label{sec:related_pcgml}

Recent machine learning methods have revolutionized the way we generate pictorial content, such as images \cite{goodfellow2014generative}. Generating content for games through machine learning is a more difficult task, however, as its representation must include the functional aspects of gameplay. Machine learning for procedural content generation (PCGML) is a fairly new family of PCG approaches \cite{summerville2018pcgml} which revolves around the use of machine learned models as a way to directly generate new content. A recent study \cite{liu2020deeplearning} surveys work on game content generation driven by deep learning specifically.

Of particular relevance to SuSketch is the work of Jain \emph{et al.} \cite{jain2016autoencoders}, in which trained auto-encoders on \textit{Super Mario Bros.} levels were used to generate new levels, as well as repair unplayable levels. The models learned the representation of the levels, and the repair process could bring problematic levels back to the manifold to ``repair'' them. Lee \emph{et al.} \cite{lee2016predicting} trained neural networks in order to predict resource locations for existing \textit{Starcraft II} (Blizzard, 2010) maps. The learned representation could also be exploited to increase or decrease the number of resources in such maps. While both projects could be used to help a designer (e.g. repairing or auto-completing their work-in-progress levels), Morai Maker \cite{guzdial2019moraimaker} is the most relevant mixed-initiative tool which takes advantage of PCGML. Morai Maker allows the user and the AI to take turns editing a platformer level for \emph{Super Mario Bros.} with PCGML detecting and enforcing local patterns between tiles. 

\subsection{Surrogate Models}\label{sec:related_surrogate}

Surrogate models (or metamodels) are data-driven function approximators that ``mimic the behavior of the simulation accurately while being computationally cheap(er) to evaluate'' \cite{gorissen2010surrogate}. Such models are expected to ``predict the  system  performance  and  develop  a  relationship between the system inputs and outputs'' \cite{gorissen2010surrogate}. Surrogate models are effective for complex simulation-based evaluations in artificial evolution \cite{togelius2011searchbased}, replacing expensive fitness evaluations with cheap fitness approximations \cite{zhou2007combining,jin2011surrogate}. Moreover, surrogate models often have a smoother fitness landscape that makes evolutionary search easier \cite{brownlee2015metaheuristic}. However, in the domain of games, academic literature is sparse. In \cite{volz2016demonstrating}, a static hand-crafted fitness function is used as a surrogate for gameplay; however, this function is not learned from observed simulation data. Morosan \emph{et al.} \cite{morosan2019automating} trained surrogate models in order to fine-tune parameters of a \emph{Ms. Pac-man} (Midway, 1982) agent, a racing car in TORCS \cite{torcs} and unit parameters in \emph{StarCraft} (Blizzard, 1998). Finally the work of Karavolos \emph{et al.} \cite{karavolos2018surrogate,karavolos2018pairing,karavolos2019multifaceted} on evolving levels and character classes via surrogate models is discussed throughout this paper and extended significantly.

\section{The Design Tool}\label{sec:tool}

SuSketch operates on a constrained level design task for a specific type of shooter game initially designed by Karavolos \emph{et al.} \cite{karavolos2018pairing,karavolos2018surrogate,karavolos2019multifaceted}, which is summarized in Section \ref{sec:tool_game}. Following interaction paradigms of past mixed-initiative tools \cite{liapis2016mixedinitiative}, SuSketch allows the designer to draw on a canvas on the left half of the screen (see Section \ref{sec:tool_interface}), while different functions, visualizations, metrics (predicted or not) and generated suggestions appear on the right half of the screen through different tabs (see Section \ref{sec:tool_analytics}).

\subsection{Design Patterns of the Target FPS Game}\label{sec:tool_game}

Through SuSketch, users design a deathmatch scenario where two players fight against each other. The winner is the player with the highest number of kills when the game ends. The game ends after 20 kills in total, or after 10 minutes. 
Players start in opposite corners of the level, and respawn there when they die; we refer to these as bases. Each player belongs to a character class throughout the game. SuSketch includes five character classes inspired by \emph{Team Fortress 2} (Valve, 2007): Heavy, Scout, Sniper, Soldier and Demolition Man. Each class has different hit points (HP), speed, and signature weapon.
The levels consist of a grid of $20\times20$ tiles, which may be impassable (very high walls) or have an elevation (distinguishing between first-floor tiles and ground-floor tiles). There are no tunnels or bridges. Stairs connect the ground floor with the first floor and players can drop from the first floor to the ground floor at any ledge. Passable tiles can contain one of three types of powerups typical of FPS games: a \emph{health pack} (increases HP up to a maximum), \emph{armor} (offers additional HP which is depleted first) and a \emph{damage boost} (player's bullets temporarily deal double damage). After some time, powerups respawn at the same location.

\begin{figure}
    \centering
    \includegraphics[width=0.95\columnwidth]{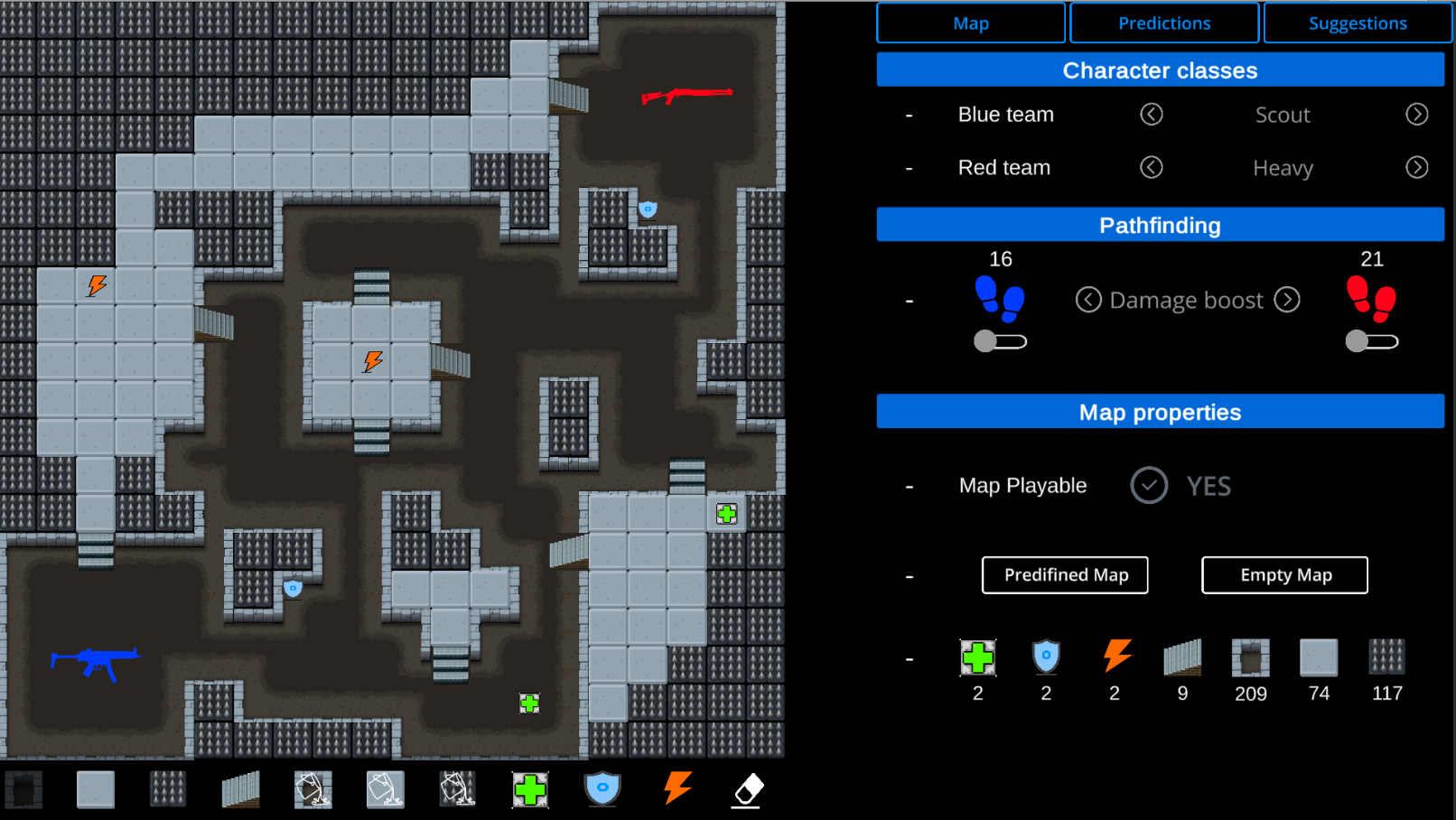}
    \caption{SuSketch drawing interface: the designer draws their desired level via the canvas and tile palette (left half) and the computer provides feedback via three tabs (right half).}
    \label{fig:interface}
\end{figure}

\subsection{Drawing Interface}\label{sec:tool_interface}

The drawing interface of SuSketch covers the left half of the screen (see Fig.~\ref{fig:interface}): it includes a grid-based view of the level, which we refer to as \emph{canvas}, and below it is a \emph{palette} for selecting which tile types to paint onto the level. Tile types include architectural features (ground-floor tiles, first-floor tiles, walls, and stairs) and powerups (damage, shield, and health). A tile can have one architectural feature and up to one powerup. Through the palette, users can also paint all tiles in a bounded area with a specific architectural feature (except stairs). The user must paint another architectural feature over an existing one; initially the level consists of ground tiles. The user can remove powerups using the eraser tool on the palette. 

\subsection{Analytics and Suggestions Tabs}\label{sec:tool_analytics}

The right half of the screen is dedicated to a set of tabs that allow the user to review the level in different ways:

\subsubsection{Map tab}\label{sec:tool_basic}

The ``default'' tab offers the necessary tools to aid the design process (see Fig.~\ref{fig:interface}). Through this tab, the user can determine the two competing players' character classes (which are also visible on the canvas as the signature weapon of the class overlayed on their base). Moreover, with each edit of the level, a number of simple metrics are computed: the number of tiles per tile type and the distance between the players' bases and other tile types. The user can choose the desired tile type, and the tool returns the average distance to all tiles of that type from each player's base. The user can view the distance (numerically and as a path on the map) from each powerup type, from stairs, and from the opponent's base.  Finally, this tab displays whether the level is well-formed and playable through a simple Yes/No notification. This sanity check is performed after every user's edit by checking whether all powerups and opponent bases are reachable (i.e. a distance can be computed) from either player's base, and by checking other constraints on architecture (e.g. stairs should be adjacent to exactly one first-floor tile). On demand, this playability check also provides details on which playability check failed; this helps the user search the level and correct the issue. Note that the other two tabs do not display any information while this playability check is failed.

\subsubsection{Predictions tab}\label{sec:tool_predicted}

\begin{figure}
    \centering
    \includegraphics[width=0.95\columnwidth]{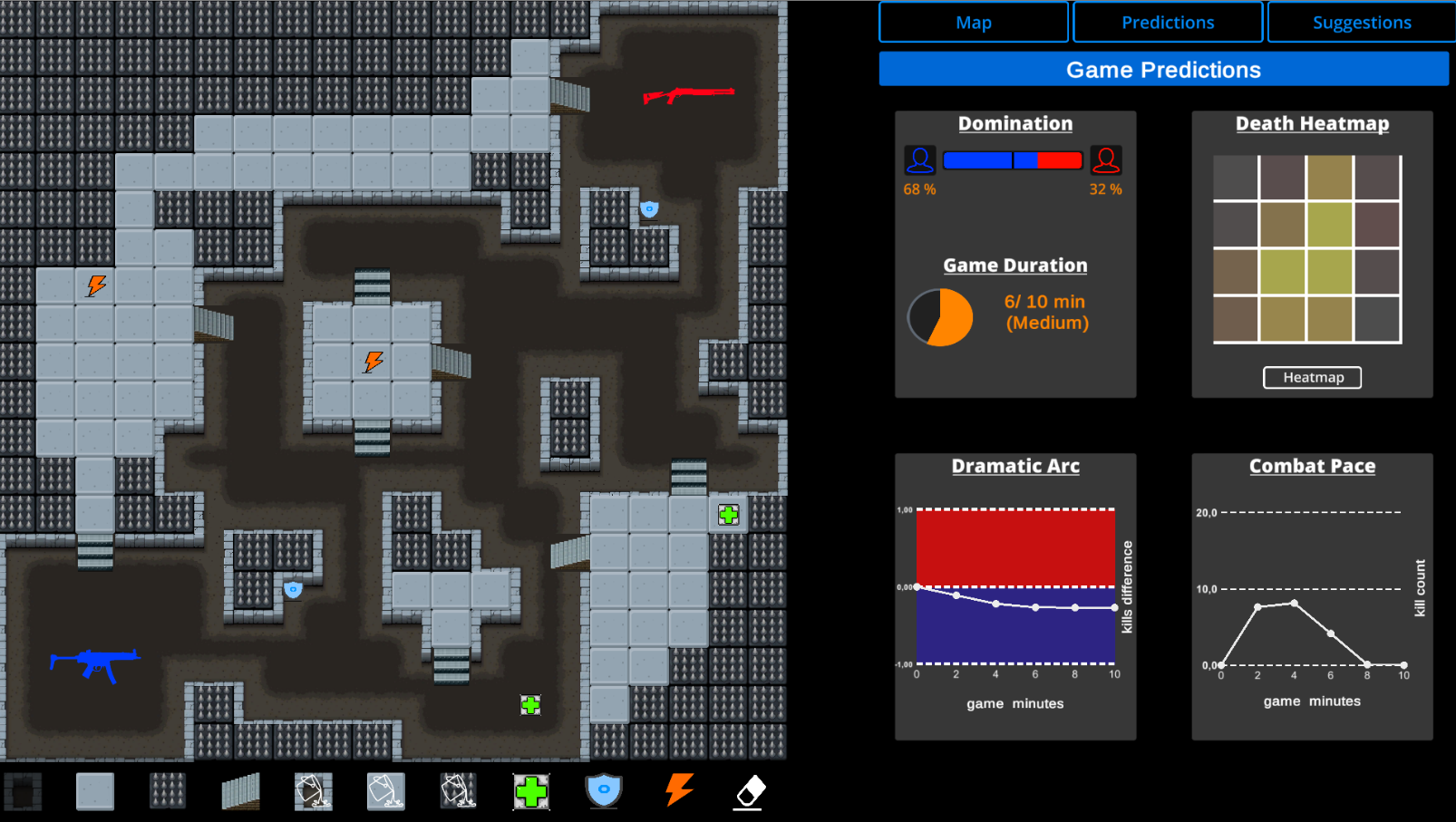}
    \caption{SuSketch gameplay predictions: the second tab shows the predicted kill ratio of players and match duration, the heatmap of predicted death locations and the temporal progression of each player's kill ratio and kills per 2 minutes.}
    \label{fig:predictions}
\end{figure}

The second tab gives the user access to a number of game analytics that constitute the core of the surrogate-based design assistant (see Fig.~\ref{fig:predictions}). The tab includes a number of metrics related to the spatial, temporal and aggregated gameplay outcomes of a hypothetical match with this level and character class setup. The aggregated metrics are the advantage of each player over the other (as the final ratio of kills per player) and the duration of the match, which were introduced in \cite{karavolos2018surrogate}. The spatial gameplay measure is a low-resolution heatmap, which shows which cells in a $4{\times}4$ grid are predicted to have more kills. The temporal gameplay metrics show the progression in terms of kills from the start of the match until the mandatory 10-minute timeout limit, in increments of 2 minutes. This progression is calculated in terms of the overall kill ratio per player up to that point (dramatic arc) and in terms of the total kills from either player during the last 2 minutes (combat pacing). The two progressions are displayed as a line plot displaying 6 data points (see Fig.~\ref{fig:predictions}). All of these metrics are calculated by  surrogate models (five networks, one per metric) trained on a large corpus of AI-produced playthroughs. Details of the training corpus, predicted metrics, network architecture, and models' accuracy are described in Section \ref{sec:surrogate}. 

\subsubsection{Suggestions tab}\label{sec:tool_suggestions}

\begin{figure}
    \centering
    \includegraphics[width=0.95\columnwidth]{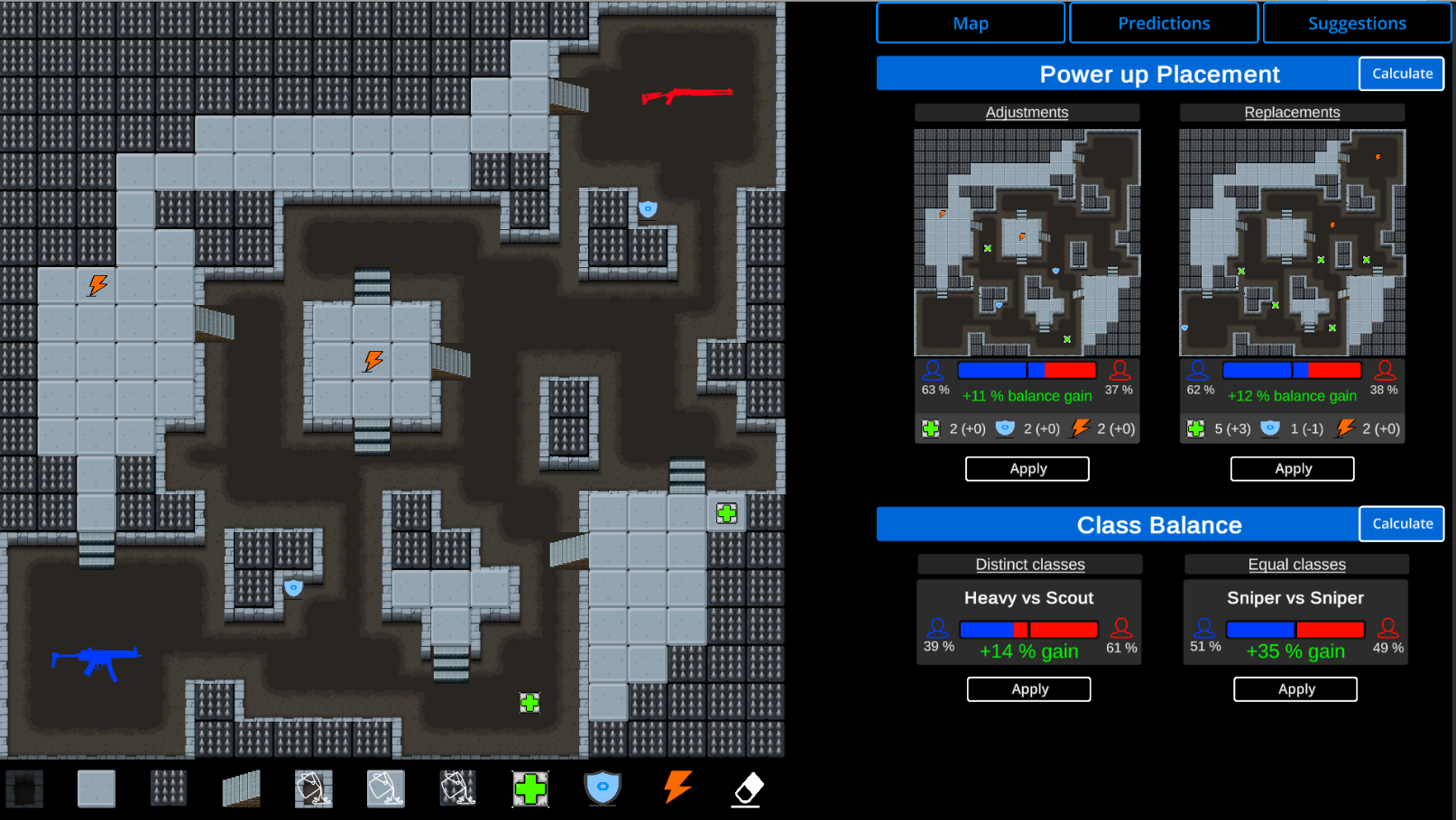}
    \caption{SuSketch suggestions: the tool visualizes two alternative powerup placements without changing the level's architecture, as well as alternative class pairings that increase the predicted balance of the matchup.}
    \label{fig:suggestions}
\end{figure}

The last tab offers AI-generated suggestions to the user (see Fig.~\ref{fig:suggestions}). While this tab is active, new suggestions are generated following every edit on the level canvas; the user can also explicitly ask for new suggestions via a `Calculate' button. Generation scripts run on separate threads and do not slow down the tool or hinder the user's design process. Two types of suggestions are currently implemented in the tool: (a) suggesting character classes for the two opposing players of the current level, and (b) suggesting new powerup placements on the current level. Both types of suggestions attempt to bring the predicted kill ratio metric closest to 50\% for a balanced match (see Section \ref{sec:suggestions} for details).

For class pairings, the tool offers two suggestions: the most balanced pair of classes when one player's class is different than the other, and the most balanced pair of classes when both players have the same class. This distinction was deemed necessary since in most cases giving both players the same class resulted in a more balanced predicted kill ratio; instead, showing two suggested class pairings may be useful to a designer who wants to diversify the playstyles. Each suggested class pair is accompanied by a visualization of the suggestion's predicted kill ratio, and a color-coded indication of the percentage improvement (in terms of kill ratio) over the current matchup. The user can replace the current class pair with the suggested one via the `Apply' button.

Level suggestions specifically target powerup placement, keeping the architecture unchanged from the user's current level. Two suggestions are offered: the first removes all powerups and places new ones randomly in the level, while the second uses a simple evolutionary strategy to adapt the current level's powerups. Details on the genetic operators and fitness calculation are provided in Section \ref{sec:suggestions}. The two level generation processes serve to (a) respect and adjust the user's powerup placement (via mutation) and (b) provide a novel powerup setup, especially when the designer has not yet added powerups to the level (best random distribution). The level suggestion interface displays a thumbnail of the generated map, as well as the numbers of each powerup type and how they differ from the user's canvas. Similar to the class suggestions, the predicted kill ratio and improvement of the suggested level is also displayed. This feedback regarding tile changes and improvement can inform a designer to accept the suggested level, replacing their canvas with the suggested level---which they can then edit further.

\section{Training Surrogate Models of Gameplay}\label{sec:surrogate}

A core innovation of SuSketch is the integration of computational models that can predict the gameplay outcomes from the level and class pairings. Earlier work by Karavolos \emph{et al.} \cite{karavolos2019multifaceted} focused on models that predict high-level aggregations of the gameplay outcome in the form of \emph{Kill Ratio} and \emph{Duration}. SuSketch includes the pre-trained computational models of these two numerical values and uses them both for visualizations in the prediction tab but also as a performance metric when generating suggestions (see Section \ref{sec:suggestions}). Since these two metrics are used with the pre-trained models described in \cite{karavolos2019multifaceted}, this Section will focus on the new gameplay outcomes introduced in SuSketch and the models trained to predict them.

\subsection{New Gameplay Evaluations}\label{sec:surrogate_evaluations}

SuSketch introduces three gameplay metrics which have multiple-value outputs:
\begin{itemize}
\item \textbf{Death Heatmap} (DH) which splits the level into $4{\times}4$ cells, and calculates the ratio of deaths by either player in this cell over all deaths. The model predicts a vector of 16 values (DH$_1$, \ldots, DH$_{16}$) within [0,1]. When showing the predictions in the tool, the outputs are normalized so that their sum equals 1.
\item \textbf{Dramatic Arc} (DA) is inspired by the notion of ``drama'' by Browne and Maire \cite{browne2010evolutionary} regarding changes in the lead during a game's progression. DA evaluates the progression of the \nth{1} player's kill ratio at a specific point in time. The model predicts a vector of 5 values (DA$_1$, \ldots, DA$_5$) representing the kill ratio of the \nth{1} player after 2, 4, 6, 8, and 10 minutes (kills are calculated cumulatively).
\item \textbf{Combat Pacing} (CP) evaluates the total number of kills scored within a time interval: it measures periods of high action. The model predicts a vector of 5 values (CP$_1$, \ldots, CP$_5$) representing the total number of kills within two minutes over the number of kills at the end of the match: for instance, CP$_1$ is the ratio of kills between 0 and 2 minutes, CP$_2$ between 2 and 4 minutes etc.
\end{itemize}
\begin{figure}
\includegraphics[width=\columnwidth]{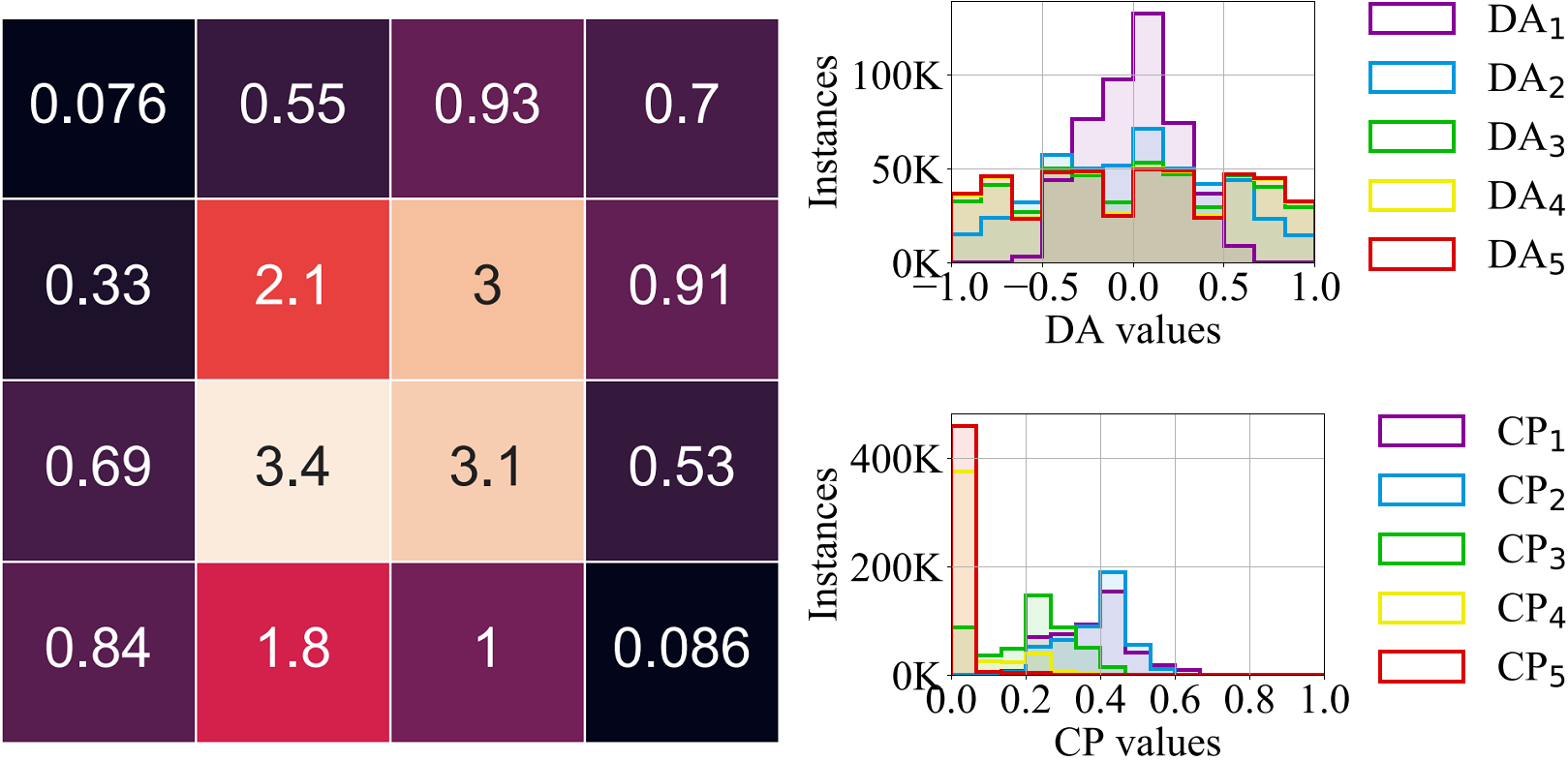}
\caption{Distribution of values in the corpus: average deaths per DH cell (left), and instances of each DA and CP value (right).}
\label{fig:dataset_distribution}
\end{figure}

\subsection{Training Corpus}\label{sec:surrogate_corpus}

Earlier work by Karavolos \emph{et al.} \cite{karavolos2018pairing} has produced a large corpus of gameplay traces performed by artificial agents on randomly generated levels and classes. In the context of this work, $4.75\cdot10^5$ gameplay traces were used to train and test the surrogate models in SuSketch. The traces contained only valid matches, i.e. matches where a total of 20 kills were scored within the 10-minute time limit. Each gameplay trace contains rich information such as the timestamp and location of a player's death, and is associated with a level layout and the parameters of the two players' character classes. 

As expected, the distribution of values for each of these measures is fairly skewed. While for single-value input such as kill ratio and duration special care was taken to split and oversample the data to avoid overfitting (see \cite{karavolos2019multifaceted} for details), there was no straightforward way to balance data containing multiple values. Figure \ref{fig:dataset_distribution} shows the average distribution of deaths in the heatmap, showing a clear tendency for deaths occurring around the center of the map. This could be due to the way the corpus was generated (since the constructive methods usually left more walls on the edges of the map) or due to the AI agents' behavioral patterns. Figure \ref{fig:dataset_distribution} also shows that DA follows a more normal distribution, but is far from the uniform distribution which would be ideal for training. The CP distribution also shows a strong tendency towards character deaths mainly occurring during the first four minutes, while the interval between 8 and 10 minutes ($CP_5$) almost never includes any player deaths.

\subsection{Machine Learning Architecture}\label{sec:surrogate_architecture}

\begin{figure*}[t]
\centering
\includegraphics[width=0.75\textwidth]{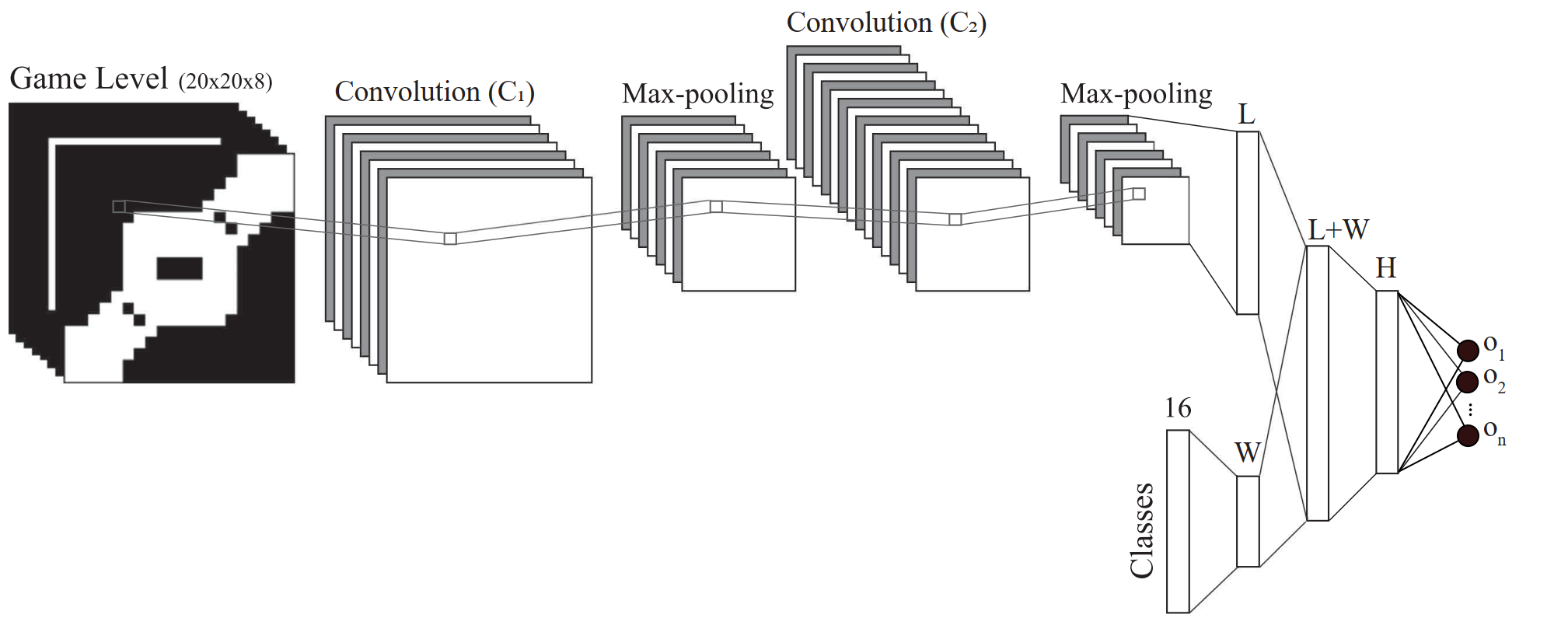}
\caption{The architecture of the CNN-based model, using the level and class parameters of both players as individual inputs. The output of the model is a $n$-dimensional vector ($n=16$ for DH, $n=5$ for CP and DA).}
\label{fig:cnn}
\end{figure*}

The input of the models is invariably the level's image, processed as a set of binary layers (one per type of tile), and the character classes' parameters normalized to [0,1] based on a pre-defined value range. The level's image is $20{\times}20$ pixels, while each character class is described by a vector of 8 parameters (hit points, speed, and 6 properties of the weapon such as rate of fire, accuracy and range). More details on the format of the input can be found in \cite{karavolos2018pairing,liapis2019fusing}.

Since every gameplay evaluation introduced in Section \ref{sec:surrogate_evaluations} consists of multiple values, the model has multiple outputs which are real values within $[0,1]$. 

Similar to earlier studies \cite{karavolos2018pairing,karavolos2018surrogate,karavolos2019multifaceted}, the level input is processed via a convolutional neural network (CNN) with two layers of convolution followed by max-pooling until a vector of real values is produced. Meanwhile, the numerical values of the character class parameters are processed through a feed forward artificial neural network (ANN), resulting in another vector of real values. The two vectors are concatenated and processed through a final ANN layer and the outputs are calculated. All hidden nodes use ReLU activation \cite{glorot2011deep}, while the output nodes use a linear function. The output of each hidden layer is normalized via batch normalization \cite{batchnorm}. The resulting architecture is shown in Figure \ref{fig:cnn}.

\subsection{Models' Accuracy}\label{sec:surrogate_experiment}

\begin{table}[b]
\centering
\caption{Best models' accuracies in the test set, averaged across all outputs of the model, and the networks' parameters. Network parameter notations are mapped on Fig.~\ref{fig:cnn}.}
\label{tab:modelling}
\small
\begin{tabular}{l|ccccc|c|c}
\textbf{Metric} & \textbf{$C_1$} & \textbf{$C_2$} & \textbf{$L$} & \textbf{$W$} & \textbf{$H$} & \textbf{$MAE$} & \textbf{$R^2$} \\
\hline\hline
DH & 32 & 64 & 64 & 16 & 128 & 0.040 & 0.400 \\ \hline
DA & 32 & 64 & 64 & 32 & 32 & 0.122 & 0.882 \\ \hline
CP & 32 & 64 & 64 & 16 & 128 & 0.042 & 0.438 \\ 
\end{tabular}
\end{table}

Table \ref{tab:modelling} shows the mean average error (MAE) and $R^2$ score for the best network architecture trained for each gameplay evaluation. Results are based on the testing accuracy from five repeated experiments with 20\% holdout (80\% of the data used for training, while reported results are on the 20\% test set), with a learning rate of 0.01 and a 10\% dropout to avoid overfitting \cite{srivastava2014dropout}. Since all models output multiple values, the MAE and $R^2$ scores are calculated on each output and averaged in Table \ref{tab:modelling}. From the results, it is obvious that DA has the highest accuracies and the model is fairly robust. Both DH and CP seem to suffer from overfitting (as a low MAE is not accompanied by a high $R^2$ score), most likely due to the skewed distribution highlighted in Section \ref{sec:surrogate_corpus}. For CP, specifically, the main reason for the low $R^2$ score was the fifth output (CP between 8 and 10 minutes) which had $R_{5}^2=0.167$; this poor performance was due to a skewed distribution of CP values at that interval, which includes many 0 values (see Fig.~\ref{fig:dataset_distribution}) as the match was already over by the \nth{8} minute.

\section{Generating Suggestions}\label{sec:suggestions}

SuSketch generates alternative class pairings and alternative powerup placements, presenting them to the user as suggestions that they can apply to replace their design. This section describes how these suggestions are generated, and evaluates their performance on a pre-set collection of 20 game levels. In all cases, the suggestions attempt to maximize the predicted balance of the matchup ($f$) which is calculated as:
\begin{equation}
f(l,c_1,c_2)=1-|2\cdot\text{KR}(l,c_1,c_2)-1|
\label{eq:balance}
\end{equation}
\noindent where KR is the predicted kill ratio for the \nth{1} player, produced by the model of \cite{karavolos2019multifaceted} with the level ($l$) and the pair of character classes ($c_1$,$c_2$) as input.

\subsection{Class Matchup}\label{sec:suggestions_class}

As discussed in Section \ref{sec:tool_suggestions}, the interface shows the most balanced class pairing when both classes are the same, and when they differ. In both cases, the system iterates over all possible combinations (5 for same-class suggestions, 20 for different-class suggestions) and chooses the one with the highest $f$ score from Eq.~\eqref{eq:balance}.

\subsection{Powerup Placement}\label{sec:suggestions_level}

Level suggestions specifically target powerup placement, keeping the architecture unchanged from the user's current canvas. As discussed in Section \ref{sec:tool_suggestions}, SuSketch uses two generation processes: \emph{adjustments} and \emph{replacements}. The replacement approach simply removes all powerups currently in the level, and iterating through each cell ($5{\times}5$ tiles) randomly allocates a powerup in a valid tile within that cell (valid tiles are not walls or stairs). This process has a 25\% chance to leave the cell empty. Only one powerup of any type can be placed in one cell in this way: the type is chosen randomly. Through this random powerup allocation, $k$ random alternatives are generated and the one with a highest $f$ is chosen.

The adjustment approach uses a simple 1+1 evolutionary strategy, mutating the current level and replacing the original level with the mutated level if its $f$ score is higher. Four different mutations are implemented: (a) moving a powerup to a random tile in the same cell (a cell is $5{\times}5$ tiles), (b) moving a powerup to a random tile in another cell than where it currently is, (c) changing a powerup's type, (d) choosing a random cell and adding a powerup if no powerup exists or removing a powerup from the cell if one exists. Except for the add/remove powerup mutation, all other mutations are applied to each powerup in the level. While in preliminary tests each mutation operator was tested on its own, in the final tool one of the above mutation operator is applied, chosen with equal probability (random mutation). After $k$ generations, the final individual is shown to the user. 

While Section \ref{sec:suggestions_experiment} explores the impact of $k$ on the fitness when adjusting or replacing powerups, in the SuSketch tool currently $k=10$ for both approaches.

\subsection{Experiments on Suggestion Generation}\label{sec:suggestions_experiment}

In order to test how the different generative approaches would perform during an actual design session, we used a broad range of levels and class pairings and compared the performance of each method in terms of final balance ($f$) compared to the original matchup's balance. 
Twenty game levels are tested, including 10 generated levels and 10 human-authored levels: details on the levels are found in \cite{karavolos2018surrogate}. In each experiment, the initial seed is one of the 20 levels using a Scout versus Heavy class matchup, as this was shown in earlier work \cite{karavolos2019multifaceted} to have the greatest potential for improvement.

\begin{figure}
\centering
\includegraphics[width=\columnwidth]{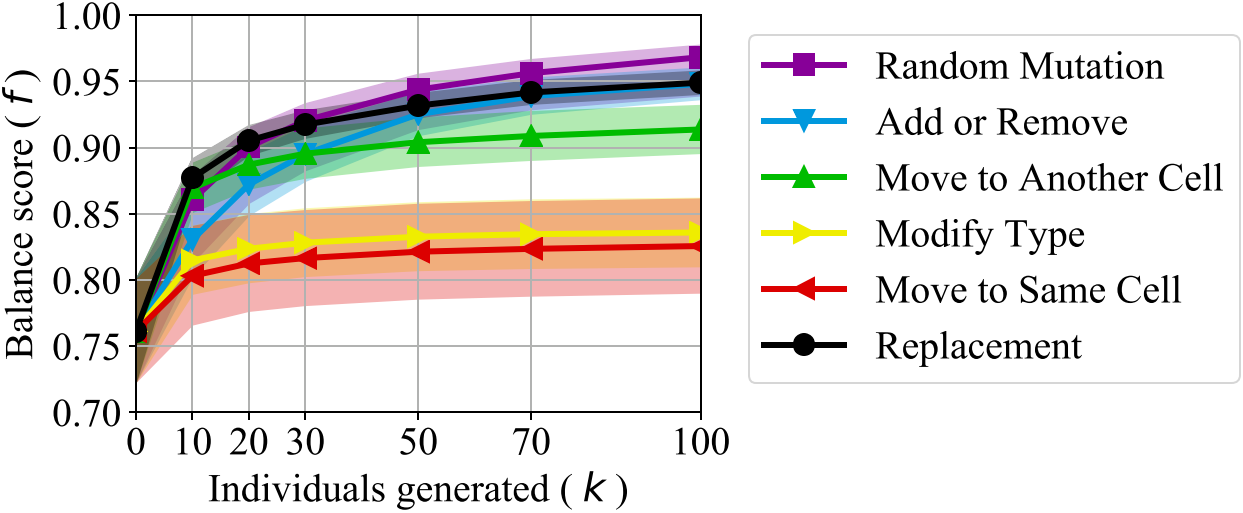}
\caption{Impact of the number of number of generated content ($k$) and the mutation types on the improvement of the games' balance. Results show the $f$ score of the fittest individual, averaged among the 20 testbed levels. The shaded areas show the 95\% confidence interval from these 20 testbed matchups.}
\label{fig:suggestions_experiment}
\end{figure}

Figure \ref{fig:suggestions_experiment} shows how  the replacement approach and the different mutation schemes for the adjustment approach affect the balance ($f$) of the 20 testbed matchups. It is evident that random mutation (which is actually used within the SuSketch tool) can improve balance fast and leads to the best performance with $k>10$. Notably, some mutations which apply small changes (e.g. move a powerup to the same cell, or change the type of powerup) lead to small improvements. Moving powerups to different cells leads to the best balance improvement after 10 generations, but seems to converge to a local optimum while mutations that apply larger changes overtake this approach given more computational resources. Interestingly, the replacement strategy which applies the most drastic changes to the level can quickly lead to balance improvements but at higher $k$ is worse than random mutation. 

Focusing on the $k=10$ benchmark, which is used within the SuSketch tool, the best percentage balance improvement from the original testbed levels is with the replacement approach (11.6\% improvement), while the adjustment approach via random mutation manages a 10\% improvement. 

In terms of the class suggestions, using the same 20 levels in a Scout versus Heavy matchup as a testbed, the average improvement for different-class suggestions was 25\%, versus 28\% for same-class suggestions. The predicted $f$ score for different-class suggestions reached $0.95\pm0.01$, versus $0.98\pm0.01$ for same-class suggestions (including the 95\% confidence intervals). The improvement with same-class suggestions is significantly higher than other possible suggestions (for classes or any level generation effort shown in Fig.~\ref{fig:suggestions_experiment}). This is not surprising, as the predictions for kill ratio (which is used to calculate $f$) mainly uses the class parameters \cite{liapis2019fusing} with limited impact from the level input. Interestingly, all best different-class suggestions were with Soldier vs. Sniper (or vice versa); the hypothesis is that both classes are long-range and the range class parameter is a strong predictor of kill ratio. The best same-class suggestions for the tested levels included all five class matchups, although Heavy vs. Heavy was only picked once while the remaining classes were evenly picked.

\section{User Study}\label{sec:userstudy}

In order to evaluate the tool's usability in real-world design settings, a user study was conducted with novice game/level designers: the protocol and results are described below.

\subsection{Experimental Protocol}\label{sec:userstudy_protocol}

The study was conducted remotely, with participants downloading a special version of SuSketch and interacting with it on their own computer. Participants had to complete a number of tasks, after which they submitted the interaction logs via e-mail to the first author and responded to an online questionnaire. Participants were recruited among the authors' game developer contacts. Therefore the research follows a blend of purposive and convenience sampling as the participants were both readily available to the researchers but also directly relevant to the task of designing game content. Notably, participants were acquaintances of the authors and thus could be positively predisposed towards the study; however, the research was carried out remotely with minimal pre- and post-study discussion with the authors to minimize any bias during the evaluation process.

The tested version of SuSketch included three tutorials, each with specific instructions. Participants were asked to complete all tutorials as well as a free-form design session. The tutorials allowed participants to use all functionalities of SuSketch, but the instructions advised them to focus on specific tasks. The first tutorial asked participants to create a level for a Heavy versus Scout (or vice versa) matchup, ensuring that (a) the level is playable, (b) the level is as close to 50\% predicted balance as possible, and (c) the predicted duration is 7 minutes or less. The instructions direct participants towards specific affordances of the interface (e.g. predictions tab, feasibility check popup). The second tutorial asked participants to create a level for Heavy versus Demolition Man, and ensure that (a) most deaths are around the center, (b) the combat pace follows a normal distribution, (c) the dramatic arc shifts from red to blue player advantage, (d) at least one suggested powerup placement is applied before the end of the session. The third tutorial asked participants to create a level for a Heavy versus Scout (or vice versa) matchup, ensuring that at least one powerup type is on the map before creating the most balanced map by repeatedly applying powerup placement suggestions. Each tutorial was only available after the previous tutorial was opened at least once. Finally, participants were asked to carry out one `free-form' design session without instructions.

The online questionnaire, which was answered after the design session was completed, collected some basic demographic information (age range, gender, level of education, game development experience, experience with game editors and experience with shooter games). The participant's e-mail address was also requested in order to cross-correlate with the interaction logs. Beyond demographics, the questionnaire included general questions based on the Post-Study System Usability Questionnaire (PSSUQ) from IBM \cite{lewis1990integrated} (16 questions in 7-item Likert scale\footnote{The only modification to PSSUQ was to reverse the scale so that 1 shows Strong Disagreement and 7 shows Strong Agreement.}), followed by 29 5-item Likert scale and 4 freeform responses on the topics of the drawing interface (11 questions), the predictions (11 questions), and the suggestions (11 questions). All questions are shown in Tables \ref{tab:pssuq_responses} and \ref{tab:appendix_responses}. The aim of the questionnaire was thus to ascertain general usability of the tool (via the PSSUQ) and the impact of the specific innovations of SuSketch on the user experience.

\subsection{Participants}\label{sec:userstudy_participants}

A total of 16 participants contributed with their interaction logs and questionnaire responses (referred below as P1-P16). Most participants identified as male (69\%), 25\% identified as female and one participant preferred not to answer. Since the participants were among the authors' contacts, it was not surprising that all participants were pursuing or had completed a Masters degree (75\%) or a Doctorate (25\%). Most participants were 25-34 years old (75\%) and the remaining were 35-55 (25\%). While most participants identified their game development role as programmers (56\%), 13\% identified as designers, 19\% as researchers and 13\% as `other'. In terms of experience with game design tools and engines, 63\% of participants self-identified as at least somewhat experienced ($\geq3$ on a 5-item Likert scale), and 81\% as at least somewhat experienced in FPS games ($\geq3$ on a 5-item Likert scale).

\subsection{Questionnaire Responses}\label{sec:userstudy_questionnaire}

\begin{table}[t]
\centering
\caption{Responses as average Likert scores on the PSSUQ part of the questionnaire, with 1 being Strong Disagreement and 7 being Strong Agreement.}
\label{tab:pssuq_responses}
\footnotesize
\begin{tabular}{p{0.76\columnwidth}|@{~}c@{~}|@{~}c@{~}|@{~}c@{~}}
\textbf{Question} & \rotatebox[origin=c]{90}{\textbf{Avg}} & \rotatebox[origin=c]{90}{\textbf{Pos}} & \rotatebox[origin=c]{90}{\textbf{Neg}} \\
\hline\hline
Overall, I am satisfied with how easy it is to use this system. &5.56 &15 &0\\ \hline
It was simple to use this system. &5.81 &15 &0\\ \hline
 I was able to complete the tasks and scenarios quickly using this system.  &5.25 &12 &1\\ \hline
I felt comfortable using this system.  &5.88 &14 &1\\ \hline
It was easy to learn to use this system. &5.94 &15 &0\\ \hline
I believe I could become productive quickly using this system.  &6.19 &15 &0\\ \hline
The system gave error messages that clearly told me how to fix problems. &5.31 &11 &2\\ \hline
Whenever I made a mistake using the system, I could recover easily and quickly.  &5.63 &13 &2\\ \hline
The information (such as online help, on-screen messages, and other documentation) provided with this system was clear.  &5.69 &12 &0\\ \hline
It was easy to find the information I needed.  &5.69 &13 &2\\ \hline
The information was effective in helping me complete the tasks and scenarios. &5.63 &14 &0\\ \hline
The organization of information on the system screens was clear.  &5.69 &13 &2\\ \hline
The interface of this system was pleasant.  &5.38 &12 &3\\ \hline
 I liked using the interface of this system.  &5.56 &12 &3\\ \hline
This system has all the functions and capabilities I expect it to have.  &5.5 &13 &1\\ \hline
Overall, I am satisfied with this system.  &5.88 &16 &0\\
\end{tabular}
\end{table}

Table \ref{tab:pssuq_responses} shows the average Likert scores from the PSSUQ, along with the positive ($>4$ on a 7-item Likert scale) and negative  ($<4$) responses. Overall, participants found the system easy and simple to use and they were overall satisfied with it. However, several participants identified that the system had a confusing interface and unclear error messages.

While some participants found information difficult to find in the general interface (including all the different tabs), all participants responded positively to the drawing interface and the ``default'' map tab; all participants found it easy to use, responsive, and were overall satisfied with it. Most participants found the path information helpful (15) and easy to understand (11). Roughly half of the participants found the display of each tile type's count helpful (8). Finally, when asked about the option to use a predefined map, most participants supported this feature, e.g.: ``I used it for trying out different classes and the dominance rate.'' (P6); ``I found the predefined map a good base to start designing my own custom map. To design a map from scratch, it would have been more difficult for me.'' (P11); other participants disagreed, e.g.:  ``The predefined map is nice but I prefer making my own map'' (P1).

Although 14 participants were satisfied with the information that the prediction panel provides, some metrics were easier to understand conceptually than others. The ranking in terms of average Likert score for individual metrics was (from easiest to understand to hardest): 1) death heatmap (16 positive responses), 2) duration, 3) combat pace, 4) domination, and 5) dramatic arc (9 positive and 2 negative responses).

In terms of the suggestions panel, 14 participants found it easy to understand. All participants found the powerup placement suggestions satisfactory in the information they provide, and 14 participants responded that they used these suggestions during their process. Participants preferred suggestions generated through the replacement approach (average Likert score: 4.75 out of 5) versus the adjustment approach (average Likert score: 4.25). On the other hand, roughly half of the participants found the class balance suggestions useful, and only 6 participants claimed that they used these suggestions (either same-class or different-class) during their process.

Summarizing the feedback of users from the questionnaire, the majority of participants could design levels using the drawing interface, even if sometimes they did not know how to react to error messages or (cryptic) feedback from the system. It is likely that the two-floor layout and many powerup types confused some participants, but the ability to load finished, predefined maps helped them understand more of the system and the map design principles. The purpose of the predicted metrics and their visualization was not always clear, especially the dominance and the dramatic arc (which are both snapshots of the same principle of player balance). The level suggestions and their visualization as a mini-map were easy to grasp and participants found themselves using them during their process. Perhaps due to the lack of clarity on the different character classes' function\footnote{SuSketch provides no information about what each name signifies, e.g. that the Scout is a fast class armed with a short-range shotgun.}, the class suggestions were not used as often.

\subsection{Interaction Logs}\label{sec:userstudy_interactions}

Observing the usage data collected from the 16 participants, a total of 55 valid\footnote{Sessions were considered ``valid'' if a map was saved at the end, if there were no technical issues with collected logs, and if more than 5 edits were made on the level canvas. 19 invalid sessions were removed (26\% of all data).} design sessions (and 55 final FPS levels) are collected. While all participants completed the \nth{1} tutorial, 1 participant did not finish the \nth{2} tutorial, 6 participants did not finish the \nth{3} tutorial, and 5 participants did not finish the free-form session. Since there were no substantial differences (e.g. learning effects) found between different tutorials in the logs, the analysis aggregates and reports findings from all logs.

The sessions were quite varied in terms of how much users interacted with the tool in general. While in 24\% of the valid sessions the participants made less than 20 edits on the level canvas (see Section \ref{sec:tool_interface}), on average 87 edits were made. In terms of the surrogate models' predictions, designers interacted with the prediction tab often, receiving on average 42 predictions per session. Note that this number is based on the number of times predictions were shown to the user while they had the predictions tab open (see Section \ref{sec:tool_predicted}) and edited the level, or when they clicked on the predictions tab. However, in 2 sessions the designers did not check the predictions tab at all. It is difficult to assess whether the number of times designers asked for predictions had an impact in their process from the logs, as it is unclear which of the predicted metrics they were trying to optimize for (if any).

In terms of the suggestions offered by SuSketch, designers interacted with the suggestions tab less often than with the predictions tab, averaging 17.6 suggestions per session. This number is based on the number of times suggestions were shown to the user while they had the suggestions tab open (see Section \ref{sec:tool_suggestions}) and edited the level, when they clicked on the suggestions tab, or when they manually asked for more suggestions via the interface. In 4 sessions the designers did not check the suggestions tab at all; actually, one designer never checked the suggestions tab during any of their sessions (3 in total). The designers used a generated suggestion to change their level or class matchup on average 5.1 times per session, although in 9 sessions no suggestions were chosen. Interestingly, in eight sessions at least 10 suggestions were chosen, with the largest number of suggestions chosen in one session being 44. In practice, therefore, it is evident that different designers found suggestions more or less useful, or were more or less open to replacing their own work with computational input. In terms of the type of suggestions chosen to replace the designers' work-in-progress, most designers chose the suggested powerup placements (4.7 times per session on average); suggested changes to class matchups were rarely chosen (0.38 times per session). Specifically, only in 13 sessions were class matchups chosen at all, and usually once (7 sessions) or twice (5 sessions). While there was no preference between class matchups (i.e. different-class pairings or same-class pairings), there seems to be a preference between powerup suggestions generated via evolutionary adjustment of the user's level (chosen 2.7 per session on average) and those generated via random replacement (2 per session) although the differences are not statistically significant (Student's $t$-test, $p>5\%$). A high positive correlation (Spearman's $\rho=0.59$) between the times a suggestion via replacement was chosen and the times a suggestion via adjustment was chosen indicates that designers who enjoyed using powerup suggestions chose between the two interchangeably.

\begin{figure}
    \centering
    \includegraphics[width=\columnwidth]{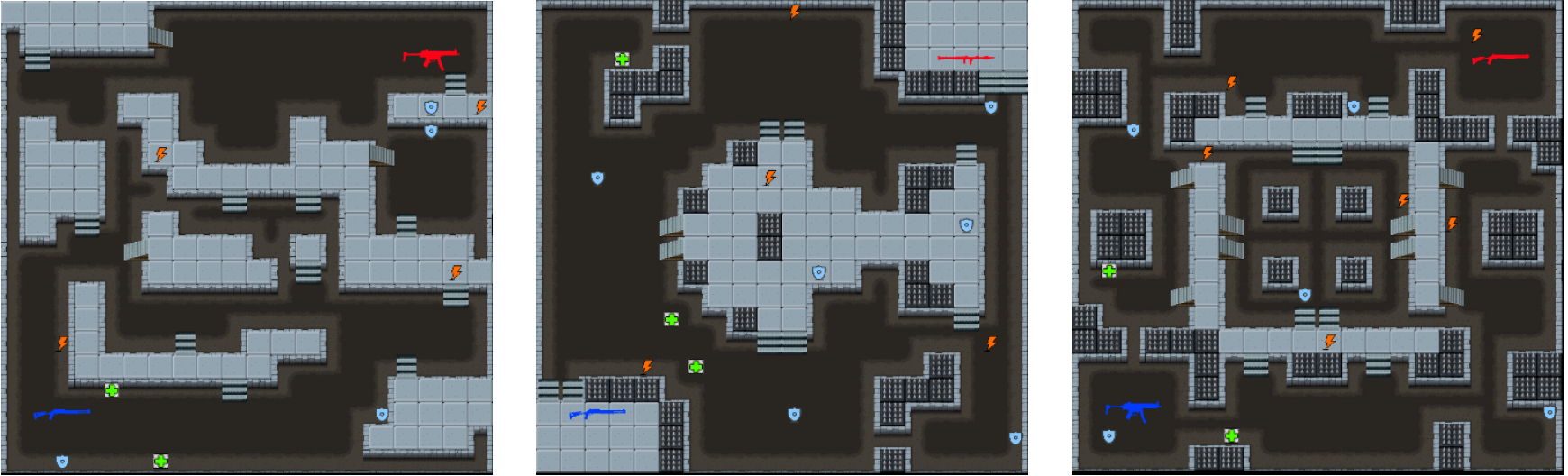}
    \caption{Sample final levels created by users.}
    \label{fig:userstudy_samples}
\end{figure}

Finally, in terms of the final FPS levels produced during the 55 design sessions, there was a broad variety of designs. In terms of architecture, most final levels had more ground floor tiles (68\% of the level's tiles, on average) than first floor tiles (20\%) or wall tiles (12\%). Interestingly, all final levels had both ground-floor and first-floor tiles, but 12 final levels had no wall tiles at all. In terms of powerups, one of the final levels did not have any powerups while on average final levels contained 12.8 powerups (out of which 4 health packs, 3.8 armors, 5 damage boosts). In 89\% of the final maps, all three types of powerups are present. Investigating the impact of suggestions, we find a positive correlation (Spearman's $\rho=0.15$) between the final number of powerups of that session and the times a suggestion via replacement was chosen, while there is no correlation ($\rho<10^{-2}$) with the number of times a suggestion via mutation was chosen. Due to the way replacements can add many powerups (unaffected by the user's own design), it is not surprising that users who tended to choose suggestions via replacement ended up with more powerups in their final design. Figure~\ref{fig:userstudy_samples} shows examples of participants' final levels.

\section{Discussion}\label{sec:extensions}

The interface design of SuSketch is aimed towards facilitating game and level design through visualizations of necessary properties (e.g. powerup counts, paths) and predicted gameplay outcomes. The responses of the user study show that designers found the more traditional aspects of the interface (canvas, pathfinding) easy to use. However, the more AI-assisted innovations (predicted metrics, suggestions) were found less easy to use and were less often used. Overall, participants in the user study possessed a high level of education and game design fluency, so it can be presumed that novices (e.g. young game designers or modders) would find some facets of SuSketch even more difficult to use. A number of extensions are envisioned for future iterations of the tool in order to address these concerns.

The current metrics displayed in the first tab of SuSketch provide the basic information which is vital for level design. Due to the interplay of first-floor tiles which can only be accessed via stairs, the paths that a player has to follow to reach points of interest are not straightforward; the shortest path visualizations allow for better feedback. However, tools such as Sentient Sketchbook \cite{liapis2013sketchbook} compute and visualize sophisticated path-based metrics such as resource safety and exploration which provide more informative quantitative measures and more granular visualizations. Calculating and visualizing such metrics would be a straightforward way to improve the feedback to the designer. Moreover, experiments in using visualizations of e.g. powerup safety for these FPS level layouts showed that more accurate predictions for gameplay metrics could be achieved \cite{liapis2019fusing}. Including this input to the deep learning models of temporal or spatial qualities could also lead to higher accuracy in these harder prediction tasks.

Currently, designers have no way of testing the game themselves, unlike Morai Maker \cite{guzdial2019moraimaker} where the designer can at any point play the work-in-progress platformer level by pressing a button. Future work should allow a similar way for users to load the 3D shooter level and play against an AI agent acting as an opponent in order to get a better feel for the matchup. However, such an implementation is more complicated and would be slower due to the need to load 3D models, the need for an AI opponent, etc. Based on participants' feedback, moreover, it is important to more clearly explain (either in text or via some visualization) the different roles and parameters that character classes have.

Several extensions are planned to enhance the suggestions generated by SuSketch. Currently level suggestions are limited to the placement of powerups, since these powerups can severely impact the balance of the game. Moreover, by respecting the human user's architectural choices, the designer is more likely to accept the proposed changes than if the tool changes in a seemingly random way the architecture that the human painstakingly defined \cite{liapis2012limitations}. However, a different set of suggestions which apply minor mutations to the architecture, as in \cite{karavolos2018pairing}, could also provide some inspiration. To better ensure that the general architectural patterns are retained, the current level symmetry could be retained with a custom embryogeny \cite{liapis2014modelingsketchbook} or by including it in the fitness calculation \cite{alvarez2018fostering}.

Regarding the suggested class pairings, the process is very simple since the possible combinations are limited. Currently SuSketch focuses on the level design task, and the five pre-authored classes makes handling classes more manageable for the user. Future iterations could allow the generator to slightly adjust the parameters of a class to customize it for a specific level. However, participants' feedback regarding class suggestions, coupled with the difficulties they had understanding the differences between classes, raises concerns whether further customizing the classes would lead to more confusion.

As a final point, the current tool provides extensive feedback to the designer in many different ways (simple path operations, predictions on gameplay, suggestions for improvement). However, better explaining the AI's decisions would help make SuSketch easier to use. Responding to a general need for explainable AI in design tools \cite{zhu2018explainable}, both the predictive models and the generated suggestions could enhance their explainability. Predictions could be explainable if e.g. an activation map \cite{selvaraju2017grad} for the current prediction is overlayed on the work-in-progress level to show the areas that need to be edited the most, similar to \cite{liapis2019fusing}. The suggestions could explain in natural language which of the powerups were changed, how (e.g. whether it was moved towards the center), and why (e.g. how much balance improves from this specific tile's change).

\section{Conclusion}\label{sec:conclusion}
This paper introduced SuSketch, a mixed-initiative tool which uses surrogate models of gameplay to both visualize difficult to simulate playthroughs and to drive the generation of suggestions for the designer to consider. While the tool builds on an extensive body of evidence and best practices in level co-creation, it offers a novel way to exploit deep learning and pattern-matching---beyond current PCGML practices---by mapping from levels and rules to gameplay outcomes. A number of extensions to the tool are also expected to increase the usability, explainability, and usefulness of suggestions.

\section*{Acknowledgments}
We would like to thank the 16 participants for their feedback and suggestions for improvements to the SuSketch tool. 
We would like to thank Daniel Karavolos for providing us with the FPS dataset and for his assistance in processing it.
This project has received funding from the EU’s Horizon 2020 programme under grant agreement No 952002.

\appendices

\section{Responses in User Questionnaire}

The additional questions that complement the PSSUQ (Table \ref{tab:pssuq_responses}) are found in Table \ref{tab:appendix_responses}, showing positive ($>3$ on a 5-item Likert scale) and negative ($<3$) responses. 

\begin{table}[hbt!]
\centering
\caption{Responses as average Likert scores (1: Strong Disagreement to 5: Strong Agreement) to the questions that followed the PSSUQ reported in Table \ref{tab:pssuq_responses}. Questions were shown in sets of 11 after the PSSUQ was completed. Questions with no Likert scores required free-form input.}
\label{tab:appendix_responses}
\footnotesize
\begin{tabular}{p{0.76\columnwidth}|@{~}c@{~}|@{~}c@{~}|@{~}c@{~}}
\textbf{Question} & \rotatebox[origin=c]{90}{\textbf{Avg}} & \rotatebox[origin=c]{90}{\textbf{Pos}} & \rotatebox[origin=c]{90}{\textbf{Neg}} \\
\hline\hline
The drawing interface was easy to use. &4.38 &14 &0\\ \hline
The drawing interface was responsive. &4.81 &16 &0\\ \hline
Overall, I am satisfied with the drawing interface.  &4.63 &16 &0\\ \hline
It was easy to use the path finding system. &4.31 &15 &0\\ \hline
The information retrieved from the path finding system was helpful. &3.94 &11 &1\\ \hline
Overall, I am satisfied with the information that the path finding system provides.  &4.00 &11 &2\\ \hline
The map playability information was easy to understand. &4.19 &12 &0\\ \hline
\multicolumn{4}{p{0.95\columnwidth}}{Describe how you have been using the predefined map button (as a base blueprint for performing additional changes, to get map ideas, etc). If you have not been using the button skip this question.}\\\hline
The information about the number of each tile and power up type was helpful. &3.75 &9 &1\\ \hline
Overall, I am satisfied with the information that the map properties panel provides. &4.00 &12 &0\\ \hline
\multicolumn{4}{p{0.95\columnwidth}}{Provide additional comments (if any) concerning the drawing interface.}\\
\hline\hline
The domination gameplay metric was easy to understand. &3.81 &10 &2\\ \hline
It was easy to use the domination gameplay metric during the map design process. &3.75 &11 &3\\ \hline
The duration gameplay metric was easy to understand. &4.63 &16 &0\\ \hline
It was easy to use the duration gameplay metric during the map design process. &4.06 &11 &1\\ \hline
The dramatic arc graph was easy to understand. &3.56 &9 &2\\ \hline
It was easy to use the dramatic arc graph during the map design process.  &3.25 &6 &3\\ \hline
The combat pace graph was easy to understand. &3.94 &11 &1\\ \hline
It was easy to use the combat pace graph during the map design process. &3.75 &12 &2\\ \hline
The death heatmap was easy to understand. &4.81 &16 &0\\ \hline
It was easy to use the death heatmap graph during the map design process. &4.31 &13 &0\\ \hline
Overall, I am satisfied with the information that the predictions panel provides. &4.25 &14 &1\\ \hline
\multicolumn{4}{p{0.95\columnwidth}}{Provide additional comments (if any) concerning the predicted gameplay metrics.}\\
\hline\hline
The interface (mini map thumbnail) of the powerup placement suggestions was easy to understand. &4.56 &14 &0\\ \hline
I was using the powerup placement suggestions during the map design process. &4.50 &14 &0\\ \hline
The suggestions provided by the adjustments minimap (first in order) assisted me in creating more balanced matchups. &4.25 &12 &2\\ \hline
The suggestions provided by the replacements minimap (second in order) assisted me  in creating more balanced matchups. &4.75 &15 &0\\ \hline
Overall, I am satisfied with the information that the powerup suggestions provide. &4.50 &16 &0\\ \hline
I was using the class balance suggestions during the map design process.  &3.27 &6 &5\\ \hline
The suggestions provided by the distinct classes tab assisted me in creating more balanced matchups. &3.47 &8 &4\\ \hline
The suggestions provided by the equal classes tab assisted me in creating more balanced matchups. &3.33 &6 &4\\ \hline
Overall, I am satisfied with the information that the class suggestions provide. &3.69 &9 &4\\ \hline
\multicolumn{4}{p{0.95\columnwidth}}{Provide additional comments (If any) concerning the generated suggestions.}
\end{tabular}
\end{table}

\bibliographystyle{IEEEtran}
\bibliography{susketch}

\begin{thebibliography}{10}
\providecommand{\url}[1]{#1}
\csname url@samestyle\endcsname
\providecommand{\newblock}{\relax}
\providecommand{\bibinfo}[2]{#2}
\providecommand{\BIBentrySTDinterwordspacing}{\spaceskip=0pt\relax}
\providecommand{\BIBentryALTinterwordstretchfactor}{4}
\providecommand{\BIBentryALTinterwordspacing}{\spaceskip=\fontdimen2\font plus
\BIBentryALTinterwordstretchfactor\fontdimen3\font minus
  \fontdimen4\font\relax}
\providecommand{\BIBforeignlanguage}[2]{{%
\expandafter\ifx\csname l@#1\endcsname\relax
\typeout{** WARNING: IEEEtran.bst: No hyphenation pattern has been}%
\typeout{** loaded for the language `#1'. Using the pattern for}%
\typeout{** the default language instead.}%
\else
\language=\csname l@#1\endcsname
\fi
#2}}
\providecommand{\BIBdecl}{\relax}
\BIBdecl

\bibitem{liapis2020tenyears}
A.~Liapis, ``10 years of the {PCG} workshop: Past and future trends,'' in
  \emph{Proc. of the FDG workshop on Procedural Content Generation}, 2020.

\bibitem{yannakakis2016panorama}
G.~N. Yannakakis and J.~Togelius, ``A panorama of artificial and computational
  intelligence in games,'' \emph{IEEE Trans. on Computational Intelligence and
  AI in Games}, vol.~7, pp. 317--335, 2015.

\bibitem{liapis2016mixedinitiative}
A.~Liapis, G.~Smith, and N.~Shaker, ``Mixed-initiative content creation,'' in
  \emph{Procedural Content Generation in Games: A Textbook and an Overview of
  Current Research}, N.~Shaker, J.~Togelius, and M.~J. Nelson, Eds.\hskip 1em
  plus 0.5em minus 0.4em\relax Springer, 2016, pp. 195--214.

\bibitem{liapis2013sketchbook}
A.~Liapis, G.~N. Yannakakis, and J.~Togelius, ``Sentient sketchbook:
  Computer-aided game level authoring,'' in \emph{Proc. of the Intl. Conf. on
  the Foundations of Digital Games}, 2013, pp. 213--220.

\bibitem{guzdial2019moraimaker}
M.~Guzdial, N.~Liao, J.~Chen, S.-Y. Chen, S.~Shah, V.~Shah, J.~Reno, G.~Smith,
  and M.~O. Riedl, ``Friend, collaborator, student, manager: How design of an
  ai-driven game level editor affects creators,'' in \emph{Proc. of the CHI
  Conf. on Human Factors in Computing Systems}, 2019.

\bibitem{novick1997mixedinitiative}
D.~Novick and S.~Sutton, ``What is mixed-initiative interaction?'' in
  \emph{Proc. of the AAAI Spring Symposium on Computational Models for Mixed
  Initiative Interaction}, 1997.

\bibitem{summerville2018pcgml}
A.~Summerville, S.~Snodgrass, M.~Guzdial, C.~Holmg{\aa}rd, A.~K. Hoover,
  A.~Isaksen, A.~Nealen, and J.~Togelius, ``Procedural content generation via
  machine learning {(PCGML)},'' \emph{IEEE Trans. on Games}, vol.~10, no.~3,
  pp. 257--270, 2018.

\bibitem{karavolos2018pairing}
D.~Karavolos, A.~Liapis, and G.~N. Yannakakis, ``Pairing character classes in a
  deathmatch shooter game via a deep-learning surrogate model,'' in \emph{Proc.
  of the FDG Workshop on Procedural Content Generation}, 2018.

\bibitem{karavolos2018surrogate}
------, ``Using a surrogate model of gameplay for automated level design,'' in
  \emph{Proc. of the IEEE Conf. on Computational Intelligence and Games}, 2018.

\bibitem{karavolos2019multifaceted}
------, ``A multi-faceted surrogate model for search-based procedural content
  generation,'' \emph{IEEE Trans. on Games}, vol.~13, no.~1, pp. 11--22, 2021.

\bibitem{shaker2013ropossum}
N.~Shaker, M.~Shaker, and J.~Togelius, ``Ropossum: An authoring tool for
  designing, optimizing and solving cut the rope levels,'' in \emph{Proc. of
  the {AAAI} Conf. on Artificial Intelligence and Interactive Digital
  Entertainment}, 2013.

\bibitem{yannakakis2014micc}
G.~N. Yannakakis, A.~Liapis, and C.~Alexopoulos, ``Mixed-initiative
  co-creativity,'' in \emph{Proc. of the Intl. Conf. on the Foundations of
  Digital Games}, 2014.

\bibitem{licklider1960mancomputer}
J.~Licklider, ``Man-computer symbiosis,'' \emph{{IRE} Trans. on Human Factors
  in Electronics}, vol.~1, pp. 4--11, 1960.

\bibitem{holland2003principles}
A.~Holland, B.~O'Callaghan, and B.~O'Sullivan, ``A constraint-aided conceptual
  design environment for autodesk inventor,'' in \emph{Principles and Practice
  of Constraint Programming}.\hskip 1em plus 0.5em minus 0.4em\relax Springer,
  2003, pp. 422--436.

\bibitem{davis2016empirically}
N.~Davis, C.-P. Hsiao, K.~Y. Singh, L.~Li, and B.~Magerko, ``Empirically
  studying participatory sense-making in abstract drawing with a co-creative
  cognitive agent,'' in \emph{Proc. of the Intl. Conf. on Intelligent User
  Interfaces}, 2016.

\bibitem{urban5}
N.~Negroponte and L.~B. Groissier, ``Urban 5: An online urban design partner,''
  \emph{IBM Report}, pp. 320--2012, Jun. 1976.

\bibitem{alvarez2018fostering}
A.~Alvarez, S.~Dahlskog, J.~Font, J.~Holmberg, C.~Nolasco, and A.~\"{O}sterman,
  ``Fostering creativity in the mixed-initiative evolutionary dungeon
  designer,'' in \emph{Proc. of the Intl. Conf. on the Foundations of Digital
  Games}, 2018.

\bibitem{smith2010tanagra}
G.~Smith, J.~Whitehead, and M.~Mateas, ``Tanagra: A mixed-initiative level
  design tool,'' in \emph{Proc. of the Intl. Conf. on the Foundations of
  Digital Games}, 2010, pp. 209--216.

\bibitem{butler2013puzzle}
E.~Butler, A.~M. Smith, Y.-E. Liu, and Z.~Popovic, ``A mixed-initiative tool
  for designing level progressions in games,'' in \emph{Proc. of the ACM
  symposium on User interface software and technology}, 2013.

\bibitem{volkovas2019mek}
R.~Volkovas, M.~Fairbank, J.~Woodward, and S.~Lucas, ``Mek: Mechanics
  prototyping tool for 2d tile-based turn-based deterministic games,'' in
  \emph{Proc. of the IEEE Conf. on Games}, 2019.

\bibitem{machado2016mechanics}
T.~Machado, I.~Bravi, Z.~Wang, A.~Nealen, and J.~Togelius, ``Shopping for game
  mechanics,'' in \emph{Proc. of the Intl. Conf. on the Foundations of Digital
  Games}, 2016.

\bibitem{perez2019gvgaibook}
D.~Perez-Liebana, S.~M. Lucas, R.~D. Gaina, J.~Togelius, A.~Khalifa, and
  J.~Liu, \emph{General Video Game Artificial Intelligence}.\hskip 1em plus
  0.5em minus 0.4em\relax Morgan \& Claypool Publishers, 2019, vol.~3, no.~2,
  \url{https://gaigresearch.github.io/gvgaibook/}.

\bibitem{machado2018ai}
T.~Machado, D.~Gopstein, A.~Nealen, O.~Nov, and J.~Togelius, ``Ai-assisted game
  debugging with cicero,'' in \emph{Proc. of the IEEE Congress on Evolutionary
  Computation}.\hskip 1em plus 0.5em minus 0.4em\relax IEEE, 2018, pp. 1--8.

\bibitem{machado2019pitako}
T.~Machado, D.~Gopstein, A.~Nealen, and J.~Togelius, ``Pitako-recommending game
  design elements in cicero,'' in \emph{Proc. of the IEEE Conf. on
  Games}.\hskip 1em plus 0.5em minus 0.4em\relax IEEE, 2019, pp. 1--8.

\bibitem{liapis2019orchestrating}
A.~Liapis, G.~N. Yannakakis, M.~J. Nelson, M.~Preuss, and R.~Bidarra,
  ``Orchestrating game generation,'' \emph{IEEE Trans. on Games}, vol.~11,
  no.~1, pp. 48--68, 2019.

\bibitem{goodfellow2014generative}
I.~Goodfellow, J.~Pouget-Abadie, M.~Mirza, B.~Xu, D.~Warde-Farley, S.~Ozair,
  A.~Courville, and Y.~Bengio, ``Generative adversarial nets,'' in
  \emph{Advances in neural information processing systems}, 2014, pp.
  2672--2680.

\bibitem{liu2020deeplearning}
J.~Liu, S.~Snodgrass, A.~Khalifa, S.~Risi, G.~N. Yannakakis, and J.~Togelius,
  ``Deep learning for procedural content generation,'' \emph{Neural Computing
  and Applications}, 2020, in Press.

\bibitem{jain2016autoencoders}
R.~Jain, A.~Isaksen, C.~Holmg{\aa}rd, and J.~Togelius, ``Autoencoders for level
  generation, repair, and recognition,'' in \emph{Proc. of the ICCC Workshop on
  Computational Creativity and Games}, 2016.

\bibitem{lee2016predicting}
S.~Lee, A.~Isaksen, C.~Holmg{\aa}rd, and J.~Togelius, ``Predicting resource
  locations in game maps using deep convolutional neural networks,'' in
  \emph{Proc. of the {AAAI} Conf. on Artificial Intelligence and Interactive
  Digital Entertainment}, 2016.

\bibitem{gorissen2010surrogate}
D.~Gorissen, I.~Couckuyt, P.~Demeester, T.~Dhaene, and K.~Crombecq, ``A
  surrogate modeling and adaptive sampling toolbox for computer based design,''
  \emph{Journal of Machine Learning Research}, vol.~11, pp. 2051--2055, 2010.

\bibitem{togelius2011searchbased}
J.~Togelius, G.~N. Yannakakis, K.~Stanley, and C.~Browne, ``Search-based
  procedural content generation: A taxonomy and survey,'' \emph{IEEE Trans. on
  Computational Intelligence and AI in Games}, vol.~3, no.~3, 2011.

\bibitem{zhou2007combining}
Z.~{Zhou}, Y.~S. {Ong}, P.~B. {Nair}, A.~J. {Keane}, and K.~Y. {Lum},
  ``Combining global and local surrogate models to accelerate evolutionary
  optimization,'' \emph{IEEE Trans. on Systems, Man, and Cybernetics}, vol.~37,
  no.~1, pp. 66--76, 2007.

\bibitem{jin2011surrogate}
Y.~Jin, ``Surrogate-assisted evolutionary computation: Recent advances and
  future challenges,'' \emph{Swarm and Evolutionary Computation}, vol.~1,
  no.~2, pp. 61--70, 2011.

\bibitem{brownlee2015metaheuristic}
A.~E. Brownlee, J.~R. Woodward, and J.~Swan, ``Metaheuristic design pattern:
  surrogate fitness functions,'' in \emph{Proc. of the Genetic and Evolutionary
  Computation Conf.}, 2015, pp. 1261--1264.

\bibitem{volz2016demonstrating}
V.~Volz, G.~Rudolph, and B.~Naujoks, ``Demonstrating the feasibility of
  automatic game balancing,'' in \emph{Proc. of the Genetic and Evolutionary
  Computation Conf.}, 2016, pp. 269--276.

\bibitem{morosan2019automating}
M.~Morosan, ``Automating game-design and game-agent balancing through
  computational intelligence,'' Ph.D. dissertation, University of Essex, 2019.

\bibitem{torcs}
B.~Wymann, E.~Espi\'{e}, C.~Guionneau, C.~Dimitrakakis, R.~Coulom, and
  A.~Sumner, ``Torcs: The open racing car simulator v1.3.7,''
  \url{http://torcs.sourceforge.net/}, 2020, accessed 22 Dec 2020.

\bibitem{browne2010evolutionary}
C.~Browne and F.~Maire, ``Evolutionary game design,'' \emph{IEEE Trans. on
  Computational Intelligence and AI in Games}, vol.~2, pp. 1 -- 16, 2010.

\bibitem{liapis2019fusing}
A.~Liapis, D.~Karavolos, K.~Makantasis, K.~Sfikas, and G.~N. Yannakakis,
  ``Fusing level and ruleset features for multimodal learning of gameplay
  outcomes,'' in \emph{Proc. of the IEEE Conf. on Games}, 2019.

\bibitem{glorot2011deep}
X.~Glorot, A.~Bordes, and Y.~Bengio, ``Deep sparse rectifier neural networks,''
  in \emph{Proc. of the Intl. Conf. on Artificial Intelligence and Statistics},
  2011.

\bibitem{batchnorm}
S.~Ioffe and C.~Szegedy, ``Batch normalization: Accelerating deep network
  training by reducing internal covariate shift,'' in \emph{Proc. of the Intl.
  Conf. on Machine Learning}, 2015.

\bibitem{srivastava2014dropout}
N.~Srivastava, G.~Hinton, A.~Krizhevsky, I.~Sutskever, and R.~Salakhutdinov,
  ``Dropout: A simple way to prevent neural networks from overfitting,''
  \emph{Journal of Machine Learning Research}, vol.~15, no.~56, pp. 1929--1958,
  2014.

\bibitem{lewis1990integrated}
J.~R. Lewis, S.~C. Henry, and R.~Mack, ``Integrated office software benchmarks:
  A case study,'' in \emph{Proc. of Human-Computer Interaction -- INTERACT},
  1990.

\bibitem{liapis2012limitations}
A.~Liapis, G.~N. Yannakakis, and J.~Togelius, ``Limitations of choice-based
  interactive evolution for game level design,'' in \emph{Proc. of AIIDE
  Workshop on Human Computation in Digital Entertainment}, 2012.

\bibitem{liapis2014modelingsketchbook}
------, ``Designer modeling for sentient sketchbook,'' in \emph{Proc. of the
  IEEE Conf. on Computational Intelligence and Games}, 2014.

\bibitem{zhu2018explainable}
J.~Zhu, A.~Liapis, S.~Risi, R.~Bidarra, and G.~M. Youngblood, ``Explainable
  {AI} for designers: A human-centered perspective on mixed-initiative
  co-creation,'' in \emph{Proc. of the IEEE Conf. on Computational Intelligence
  and Games}, 2018.

\bibitem{selvaraju2017grad}
R.~R. Selvaraju, M.~Cogswell, A.~Das, R.~Vedantam, D.~Parikh, and D.~Batra,
  ``Grad-cam: Visual explanations from deep networks via gradient-based
  localization,'' in \emph{Proc. of the IEEE Conf. on Computer Vision}, 2017,
  pp. 618--626.

\end{thebibliography}

\end{document}